\documentclass[preprint,12pt]{elsarticle}
\usepackage{graphicx}
\usepackage{amssymb}
\usepackage{amsthm}
\usepackage{amsmath}
\usepackage{xcolor}
\usepackage{booktabs}
\usepackage{multirow}
\usepackage{url}
\usepackage{array}              
\usepackage{multirow}           
\usepackage{makecell}           
\usepackage{amssymb}            
\usepackage{pifont}             
\usepackage{array}      
\usepackage{multirow}   
\usepackage{makecell}   
\usepackage{amssymb}    
\usepackage{booktabs}   
\usepackage{xcolor}     
\usepackage{graphicx}   
\usepackage{float}
\usepackage{algorithm}
\usepackage{algpseudocode}
\usepackage{amsmath}
\usepackage{enumitem}
\usepackage{tikz}
\usepackage{pgf-pie}
\usepackage{subcaption}
\usepackage{xcolor}

\definecolor{colComposite}{RGB}{189,189,189}
\definecolor{colBuckling} {RGB}{215, 48, 39}
\definecolor{colNonlinear}{RGB}{140, 81,200}
\definecolor{colStatics}  {RGB}{253,174, 97}
\definecolor{colDynamics} {RGB}{ 26,152, 80}

\usepackage{tcolorbox}
\usepackage{fvextra}         
\usepackage{ragged2e}
\usepackage{microtype}
\tcbuselibrary{skins, breakable, hooks}

\newtcolorbox{systempromptbox}[1]{
  enhanced,
  breakable,
  colback=white,
  colframe=colNonlinear,
  colbacktitle=colNonlinear,
  coltitle=white,
  fonttitle=\bfseries\small,
  title={#1},
  arc=3pt,
  boxrule=1pt,
  top=3mm, left=4mm, right=4mm, bottom=3mm,
}

\newtcolorbox{userpromptbox}[1]{
  enhanced,
  breakable,
  colback=white,
  colframe=colDynamics,
  colbacktitle=colDynamics,
  coltitle=white,
  fonttitle=\bfseries\small,
  title={#1},
  arc=3pt,
  boxrule=1pt,
  top=3mm, left=6mm, right=6mm, bottom=3mm,
  before upper={\ttfamily\justifying},
}

\newtcolorbox{userrequirementbox}[1]{
  enhanced,
  breakable,
  colback=white,
  colframe=colStatics,
  colbacktitle=colStatics,
  coltitle=white,
  fonttitle=\bfseries\small,
  title={#1},
  arc=3pt,
  boxrule=1pt,
  top=3mm, left=6mm, right=6mm, bottom=3mm,
  before upper={\justifying},
}
\newcommand{\cmark}{\textcolor{green}{\ding{51}}}
\newcommand{\xmark}{\textcolor{red}{\ding{55}}}

\renewcommand{\arraystretch}{1.30}

\journal{}

\begin{document}

\begin{frontmatter}


\title{A Multi-AI-agent Framework Enabling End-to-end Finite Element Analysis for Solid Mechanics Problems}



\author[add1]{Titu Ranjan Sarker}
\author[add1]{Muhammed Jawaad Zulqernine}
\author[add3]{Ling Yue}
\author[add3]{Shaowu Pan}
\author[add1]{Chenxi Wang}
\author[add1,add2]{Shiyao Lin\corref{cor1}}

\address[add1]{University of Texas at Arlington, Arlington, TX 76010, USA}
\address[add2]{Institute for Predictive Performance Methodologies, Fort Worth, TX 76118, USA}
\address[add3]{Rensselaer Polytechnic Institute, Troy, NY 12180, USA}

\cortext[cor1]{Corresponding author: Shiyao Lin, shiyao.lin@uta.edu}

\begin{abstract}
Finite element analysis (FEA) is the most important numerical approach for solid mechanics. Challenges of FEA include a steep learning curve for entry-level users and potential false simulations due to incorrect definitions of key simulation components, such as boundary conditions, load cases, and solution variables. Years of engineering experience are usually necessary for real-world problem-solving. To address these issues, we present AbaqusAgent, a multi-agent framework grounded in large language models (LLMs) for solid mechanics analyses. AbaqusAgent is developed to facilitate analysis case generation and execution using Abaqus, one of the most widely used FEA packages, by turning users’ natural-language instructions into executed FEA analyses and result visualization. AbaqusAgent is composed of six agents, including interpreter, architect, input writer, runner, reviewer, and visualizer agents, encompassing all the essential pre-processing and post-processing steps of standard FEA analyses. A wide variety of 50 solid mechanics problems have been successfully validated, achieving an overall success rate of 86\%. Beyond improving the efficiency of FEA for solid mechanics problems and lowering the barrier to computational mechanics education,  AbaqusAgent advances the human-simulation interaction paradigm and enables integration with AI-empowered optimization and material characterization workflows. The code is available at:\url{https://github.com/LIRAM-LIN/AbaqusAgent}

\end{abstract}


\end{frontmatter}


\section{Introduction}
\label{secIntro}

In recent years, the fast development of large language models (LLMs) and other generative AI models has rapidly enabled the application of AI agents in many scientific disciplines. AI agents have streamlined complex workflows to achieve end-to-end scientific procedures. For example, computational fluid dynamics (CFD) has been largely automated with LLM-enabled agents, where users' prompts can be turned into simulated fluid flow \citep{foamAgent,metaFoam,foamgpt, LLMFLUID}. In bioengineering, gene-editing has been automated with AI agents for selecting systems, planning experiments, choosing methods, drafting protocols, designing assays, and analyzing data \citep{CRISPRGPT}. A system of multi-modal AI agents integrated with extended-reality equipment has been developed for assisting lab operations in biomedical laboratory settings \citep{labOS}.  In chemistry, an LLM-based reaction development framework (LLM-RDF) has been developed to handle fundamental tasks in chemical synthesis development \citep{chemcrew}.  The LLM-RDF is composed of six pre-prompted agents, allowing chemists to leverage the automated experimental platforms for various synthesis tasks \citep{chemcrew}. El Agente, an LLM-powered multi-agent system, was developed to execute quantum chemistry computations based on natural-language instructions, making advanced quantum chemistry accessible to both experts and non-specialists \citep{ElAgente}. In material science, a closed-loop, agentic LLM workflow that autonomously extracts thermoelectric and structural properties from about 10,000 full-text scientific articles, enabling large-scale, verifiable data extraction from
unstructured scientific literature \citep{AIAgentMaterialProperty}. The potential of LLM innovations for experimental methodologies, data analysis, and knowledge extraction within the electronic laboratory notebook (ELN) framework to advance material science research was explored in \citep{LLMMaterialScience}. A hackathon was organized to explore the possibilities of applying LLMs in chemistry and materials science \citep{llmhackathon}. Applications, including property predictions for molecules and materials, material interface design, knowledge extraction, and education module development, were reported. The applications of agentic AI in science are currently speeding up and reshaping the research workflows in many domains \citep{AI4Science1, AI4Science2, AI4Science3,AI4Science4}.

\textbf{Related work on AI-agent-enabled finite element analysis (FEA).} FEA is one of the most important analysis methods in almost all the disciplines of modern engineering. Originating in the 1950s \citep{FEA1},  FEA has evolved into the mainstream method for validating preliminary designs, reducing experimental costs, and enabling novel digital twins in the modern engineering ecosystem. However, applying FEA for real-world problems necessitates in-depth knowledge of interdisciplinary science and significant engineering experience. Careless or “black-box” usages of FEA may lead to catastrophic consequences \citep{FEAEdu}. 

Conventional automated FEA workflows are primarily based on application programming interfaces (APIs) of FEA packages. Abaqus-Python API has been utilized for performing parametric studies related to structural safety \citep{abaqusAPI1}, conducting general 3D topology optimization \citep{abaqusAPI2}, generate complex geometries \citep{abaqusAPI6,abaqusAPI7,abaqusAPI8}, and investigating progressive damage behaviors in aerospace materials \citep{abaqusAPI3,abaqusAPI4,abaqusAPI5}. However, the API-based FEA automation often lacks flexibility as it depends on predefined scripts, fixed modeling templates, and solver-specific parameters. New geometry, boundary conditions, and solver setups usually require modifications or recreations of the existing API scripting templates.

Recent developments of LLMs can potentially mitigate the entry difficulties for FEA users by translating natural-language engineering instructions into structured simulation workflows, thus reducing the need for extensive prior expertise in pre-processing (e.g., 3D modeling, meshing, boundary condition assigning, etc.), solver setup, and postprocessing. Hou et al. integrated LLMs
with FEA to automate the generation of input files of an FEA software, Calculix \citep{calculix}. The approach utilized a planning method and a graph convolutional network (GCN)-transformer model to enhance the accuracy and reliability of the generated simulations \citep{autoFEA}. An LLM-based interactive system was developed to streamline the operation flow of computer-aided design (CAD) and FEA software to improve the efficiency connecting engineering design and problem-solving \citep{CADFEA}. An LLM-based agent, ModSolAgent, was designed to automatically generate Abaqus Python scripts for FEA modeling. In ModSolAgent, a distilled Abaqus instruction–code dataset was used to fine-tune lightweight open-source LLMs for the Python scripting tasks \citep{ModSolAgent}. An automated system, MooseAgent \citep{mooseAgent}, for the Moose multiphysics simulation framework \citep{moose} was developed to interpret simulation requirements from natural language to automatically generate Moose input files. Qi et al. \citep{feaGPT} developed an agentic-AI framework for the FEA of turbocharger and aircraft wings (NACA4412 wing structure) cases to streamline the parametric studies. An LLM-empowered computer-aided engineering (CAE) agent was developed to automate mathematical derivations and implementation. The LLM-enabled model order reduction (MOR) can effectively reduce the computational cost of simulations \citep{llmCAE}.  A multi-agent framework, named MechAgents, was developed for solid mechanics problems based on an open-source FEA package FEniCS \citep{MechAgent}.  MechAgents can plan, formulate the mechanics problem, write and debug FEniCS scripts, execute simulations, and critique results. Following \citep{MechAgent}, the collaboration dynamics and reliability challenges of multi-agent LLM systems were characterized in \citep{CollaborationDynamics}. AutoGen-based \citep{autogen} agentic frameworks for FEA problems were tested in \citep{MechAgent} to evaluate seven role configurations across four tasks. How agentic role composition and the interaction patterns affect the reliability of AI-agent-enabled FEA was characterized. \citep{CollaborationDynamics}. Fine-tuned LLMs with multi-agent systems were developed for automated FEA in \citep{ALLFEM}. The work was motivated by the findings that off-the-shelf LLMs are usually not reliable for processing specific FEA tasks. Domain-specific fine-tuning of LLMs was performed. The highest success rate, 71.79\%, was achieved with a fine-tuned GPT-OSS 120B and a multi-agent framework on a 39-problem benchmark.

\begin{table}[]
  \caption{Comparison of existing FEA AI agents and this work.}
  \label{AgentComparisons}
\centering
\resizebox{\textwidth}{!}{%
\begin{tabular}{llll}
\hline
\textbf{AI Agent} & 
\textbf{FEA Solver}   & \textbf{LLM Used}                      & \textbf{Validation}\\
\hline
MechAgents \citep{MechAgent}& FEniCS   & GPT-4                       & Two successful stress analyses                                                          \\
ALL-FEM \citep{ALLFEM} &
  FEniCS &
  \begin{tabular}[c]{@{}l@{}}Llama 3.2, Qwen3, GPT-5 \end{tabular} &
  \begin{tabular}[c]{@{}l@{}}71.79\% success across 39 cases\end{tabular} \\
MooseAgent \citep{mooseAgent} & MOOSE    & DeepSeek-R1 and -V3 & 93\% success across 9 cases\\
This work & Abaqus    &Opus 4.6 & 86\% success across 50 cases  \\
\hline
\end{tabular}}
\end{table}

\textbf{Focus of this work.} AbaqusAgent in this paper aims to harness AI agents to run FEA simulations automatically, enabling the potential to benefit a large user group for engineering problem-solving and education. This work focuses on three major aspects of AbaqusAgent, including (1) the architecture and orchestration of the agents (see Section \ref{secAgentFramework}), (2) the curation of a versatile RAG database for Abaqus benchmark problems and scripts (see Section \ref{secAgentFramework}), and (3) the accuracy, efficiency, and LLM token usage for solving versatile solid mechanics problems (see section \ref{secResults}). The overall performance of AbaqusAgent is compared with relevant works in Table \ref{AgentComparisons}.

\textbf{Main contributions.} The main contributions in this paper include:

\begin{itemize}
\item We have developed a six-agent architecture. With the orchestrated agents, AbaqusAgent is capable of solving a wide variety of 50 solid mechanics problems with 86\% success rate. 
\item {We have curated a heterogeneous and inclusive RAG repository of 104 solid mechanics problems. Along with the RAG, we developed a hierarchical scoring-based searching strategy to efficiently and effectively search for the most similar curated case and template Abaqus input file for the LLM to follow and modify.} 
\item We have performed a systematic ablation study on the effectiveness of agent architecture, RAG, self-correction, and prompt quality.  
\item AbaqusAgent targets Abaqus, the dominant FEA platform in industry, chosen for its capabilities for complex problems and broad deployment across critical domains, ensuring the framework is immediately deployable within existing infrastructure and maximizing real-world impact. 
\end{itemize}
\section{Methodology}
\label{secAgentFramework}

\subsection{Agent components and orchestration}
\begin{figure}[!htbp]
  \centering
  \includegraphics[scale=0.4]{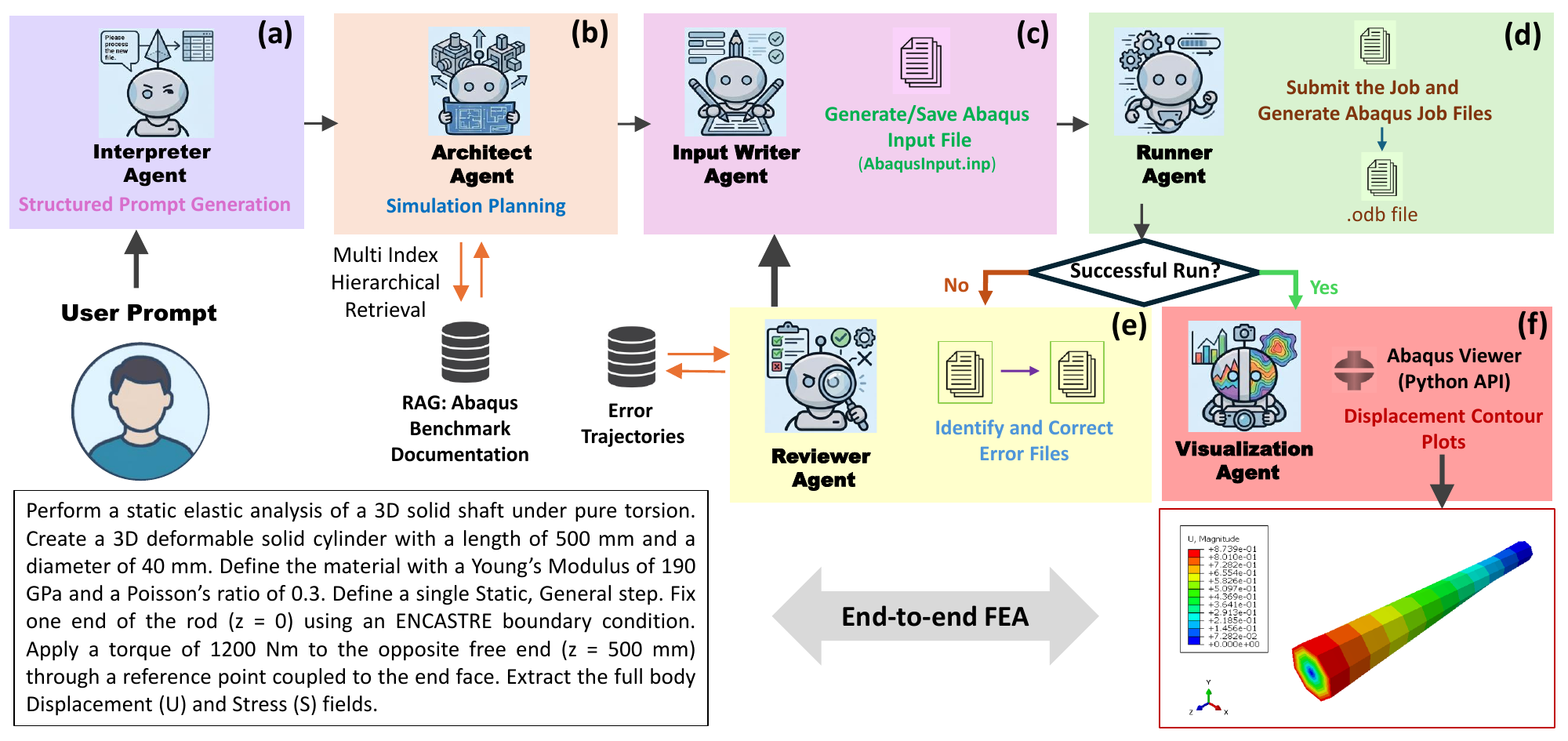}
  \caption{The AbaqusAgent architecture enables an end-to-end workflow that converts natural language input into post-processing visualizations. It includes six agents: the \textbf{Interpreter Agent}, which validates requirements and refines into structured prompts; the \textbf{Architect Agent}, which retrieves similar Abaqus cases; the \textbf{Input Writer Agent}, which generates FEA scripts; the \textbf{Runner Agent}, which executes the scripts and produces FEA results; the \textbf{Reviewer Agent}, which identifies and corrects running errors; and the \textbf{Visualization Agent}, which processes the FEA results exports deformation data and contours as PNG images and CSV files. (Agent icons generated using Gemini)}
  \label{fig_agentArchitecture}
\end{figure}

AbaqusAgent is organized as a modular system consisting of six core agents as illustrated in Figure \ref{fig_agentArchitecture}. This multi-agent framework processes natural-language requirements, uses RAG to identify analogous cases, constructs the simulation setup, executes the analysis, diagnoses errors, iteratively applies corrections, and visualizes the simulation results. 

\textbf{Interpreter Agent:} The Interpreter Agent is the first node of the AbaqusAgent. Its main role is to evaluate user requirements and identify five key parameters for Abaqus modeling: (1) geometry details, (2) material properties, (3) boundary conditions, (4) loading conditions, and (5) results output request. If any parameter is missing, the agent notifies the user and requests additional information. The node also paraphrases the prompt to ensure the Architect Agent can overcome information insufficiency and misinformation due to low user prompt quality.

\textbf{Architect Agent:}
The Architect Agent operates after the Interpreter Agent. It parses user requirements into four elements: (1) case name, (2) domain, (3) category, and (4) material. The Architect Agent uses LLM to generate and parse these components according to user input. With this information, the Architect Agent performs a similarity search, followed by a secondary hybrid search (Algorithm~\ref{alg:similarity-search}). Facebook AI Similarity Search (FAISS) \citep{FAISS} is first used to search 104 documented cases in the RAG repository, selecting similar cases by case name and filtering by domain. The secondary filter applies weighted searching using case name, category, and material to identify the most similar case. Through multiple trials, the optimal weights for case name, category, and material were set at 30\%, 60\%, and 10\%, respectively, to maximize search accuracy. Once the most similar case is selected, the task proceeds to the Input Writer Agent.

\textbf{Input Writer Agent:} The Input Writer Agent creates a new Abaqus input file based on user requirements. The created Abaqus input file defines the simulation model and provides instructions for the Abaqus solver to run the FEA simulation. The input file retrieved by the Architect Agent serves as a template, with additional formatting and modeling constraints provided by the system prompt. The generated input file follows the required Abaqus structure, including part definition, material assignment, assembly, step definition, boundary conditions, loading, mesh generation, and job setup. The completed input file is then passed to the Runner Agent for execution.

\begin{algorithm}[!htbp]
\caption{Hybrid Similarity Search}
\label{alg:similarity-search} 
\begin{algorithmic}[1]
\Require Query $q$, weights $w = \{w_{cat}=0.30,\ w_{name}=0.60,\ w_{mat}=0.10\}$
\Ensure Most similar case $c^*$
\State \textbf{// Stage 1: Semantic Retrieval}
\State $(q_{cat}, q_{name}, q_{mat}, d) \leftarrow \text{ParseQuery}(q)$ \Comment{Extract structured fields: category, name, material, domain}
\State $E \leftarrow \text{Embed}(q_{cat}, q_{name}, q_{mat}, d)$ \Comment{Embed structured fields as query}
\State $R_i \leftarrow \text{TopK}(E, k=15)$ \Comment{Retrieve top-k matches}
\State \textbf{// Stage 2: Hard Filter and Weighted Ranking}
\State $R_d \leftarrow \{x \in R_i \mid \text{domain}(x) = d\}$ \Comment{Hard filter by case domain}
\If{$R_d = \emptyset$}
    \State $R_d \leftarrow R_i$ \Comment{Fallback to all candidates if no domain match}
\EndIf
\For{each case $x \in R_d$}
    \State $\text{score}(x) \leftarrow w_{cat} \cdot \text{sim}(x_{cat}, q_{cat}) + w_{name} \cdot \text{sim}(x_{name}, q_{name}) + w_{mat} \cdot \text{sim}(x_{mat}, q_{mat})$
    \Comment{Weighted priority: 30\% category, 60\% name, 10\% material}
\EndFor
\State $R_{ranked} \leftarrow \text{SortDescending}(R_d,\ \text{score},\ \text{faiss\_score})$ \Comment{Sort by hybrid score, FAISS score as tie-breaker}
\State $c^* \leftarrow R_{ranked}[1]$ \Comment{Select highest scoring case}
\State \Return $c^*$
\end{algorithmic}
\end{algorithm}

\textbf{Runner Agent:} The Runner Agent executes the Abaqus input file and generates the result files, including the output (.odb) file, for post-processing. Upon completion, the agent forwards the .odb file to the Visualization Agent. If the simulation fails, it sends the debug files to the Reviewer Agent.

\textbf{Reviewer Agent:} The Reviewer Agent analyzes error files and proposes corrections using the LLM’s reasoning capabilities and guidance from an Abaqus expert’s system prompt. Corrected error information is sent to the Input Writer Node, and this process repeats, as shown in Algorithm~\ref{alg:iterative-refinement}, until the simulation succeeds. The maximum number of iterations for all tests is 15. Cases not resolved within 15 iterations are considered unsolved. Although some complex problems may require up to 40 iterations, the limit is set at 15 to optimize time and cost.

\begin{algorithm}[!htbp]
\caption{Iterative Refinement Process}
\label{alg:iterative-refinement}
\begin{algorithmic}[1]
\Require Natural language requirement $R$, maximum iterations $M$
\Ensure Final simulation configuration $F^*$
\State $P \leftarrow \text{Interpreter}(R)$ \Comment{Interpret the user requirement}
\State $G \leftarrow \text{Architect}(P)$ \Comment{Retrieve structural template and generate initial plan}
\State $F^0 \leftarrow \text{Input Writer}(P, G)$ \Comment{Initial input file generation using retrieved template}
\State $H \leftarrow \{\}$ \Comment{Initialize history}
\For{$i \leftarrow 1$ \textbf{to} $M$}
    \State $L^i, S^i \leftarrow \text{Runner}(F^{i-1})$ \Comment{Execute simulation}
    \If{$S^i = \text{SUCCESS}$}
        \State \Return $F^{i-1}$ \Comment{Simulation successful}
    \EndIf
    \State $H \leftarrow H \cup \{(F^{i-1}, L^i)\}$ \Comment{Update history}
    \State $E^i \leftarrow \text{ParseErrors}(L^i)$ \Comment{Extract errors}
    \State $\Delta F^i \leftarrow \text{Reviewer}(E^i, F^{i-1}, H)$ \Comment{Generate corrections}
    \State $F^i \leftarrow \text{Input Writer}(P, G, F^{i-1}, \Delta F^i)$ \Comment{Rewrite input file using reviewer corrections}
\EndFor
\State \Return $F^M$ \Comment{Return best attempt}
\end{algorithmic}
\end{algorithm}

\textbf{Visualization Agent:} The Visualization Agent uses python scripting to open the successful completed Abaqus analyses for detailed result visualization and automatically saves deformation values and contour plots of the analyzed object.

\subsection{Retrieval-augmented generation (RAG)}
As a closed-source commercial package, the benchmark documentation and scripts of Abaqus are not as readily available as those for open-source FEA codes, such as MOOSE, Caculix, and FEniCS \citep{moose,calculix,fenics}.  The curation of the RAG for AbaqusAgent poses a more significant challenge due to the limited resources. In this paper, we curated an RAG database consisting of 104 solid mechanics problems. Among the 104 problems, 71 were from the Abaqus benchmark manual and 33 textbook-style problems were developed by us. 

AbaqusAgent uses a structured retrieval strategy with a single FAISS vector store, as outlined in Algorithm~\ref{alg:similarity-search}. Each case is represented by structured fields: case name, analysis domain, category, and material type. In retrieval, FAISS returns candidate cases based on semantic similarity. These candidates are filtered by exact analysis-domain match and re-ranked using a weighted hybrid score that considers case name, category, and material similarity. This approach increases retrieval precision and ensures physical consistency across Abaqus structural analysis tasks.

\textbf{Documented Abaqus case structure:} The knowledge base is developed by systematically parsing a curated set of Abaqus benchmark and tutorial cases from diverse analysis domains and structural configurations. Information is organized along three complementary dimensions to capture both the physical intent and technical implementation of each reference case.

\textit{Case metadata (the first dimension)}, defines each simulation case using four attributes: case name, analysis domain, physical category, and material classification. The analysis domain distinguishes among simulation families, including static, buckling, dynamic, frequency, and vibration analyses. The physical category specifies the structural context, identifying configurations such as beams under axial loading, plates under transverse pressure, and shells under thermal loading. The material field classifies the model's constitutive nature, such as metals, polymers, and other engineering materials. These attributes form the primary filtering layer of the retrieval system, ensuring candidate cases are physically aligned with the user prompt before further comparison.

\textit{Problem description (the second dimension)}, offers a natural-language summary of each case, which provides details about the structural geometry, loading conditions, boundary constraints, material behaviors, and analysis objectives. The problem description supports semantic matching, enabling the retrieval system to identify conceptually similar references even when terminology varies.

\textit{Input file contents (the third dimension)}, includes the complete Abaqus input syntax for each reference case. The input files are either from Abaqus user documentation or validated curation. This dimension serves as the primary reference for the Input Writer Agent, providing syntactically correct and contextually validated templates for all supported analysis families.

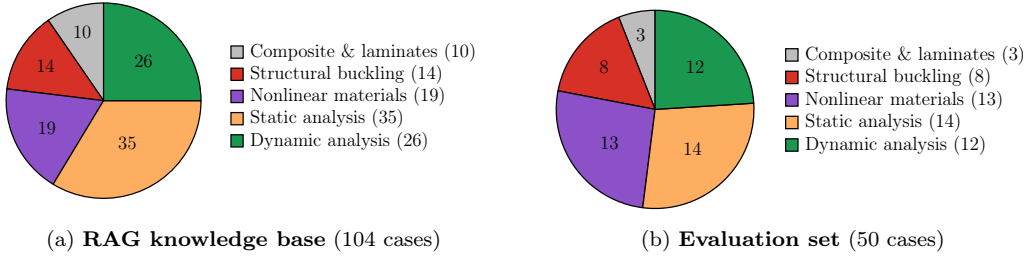
\begin{figure}[!htbp]
    \centering
    \begin{subfigure}[t]{0.47\textwidth}
        \centering
        \resizebox{\textwidth}{!}{%
        \begin{tikzpicture}
            \pie[
                radius      = 2.2,
                color       = {colComposite, colBuckling,
                               colNonlinear, colStatics, colDynamics},
                text        = legend,
                before number = ,
                after number  = ,
                every label/.style = {font=\small},
                sum         = auto,
                rotate      = 90
            ]{
                10/Composite \& laminates (10),
                14/Structural buckling (14),
                19/Nonlinear materials (19),
                35/Static analysis (35),
                26/Dynamic analysis (26)
            }
        \end{tikzpicture}
        }
        \caption{\textbf{RAG knowledge base} (104 cases)}
        \label{fig:pie_rag}
    \end{subfigure}
    \hfill
    \begin{subfigure}[t]{0.47\textwidth}
        \centering
        \resizebox{\textwidth}{!}{%
        \begin{tikzpicture}
            \pie[
                radius      = 2.2,
                color       = {colComposite, colBuckling,
                               colNonlinear, colStatics, colDynamics},
                text        = legend,
                before number = ,
                after number  = ,
                every label/.style = {font=\small},
                sum         = auto,
                rotate      = 90
            ]{
                3/Composite \& laminates (3),
                8/Structural buckling (8),
                13/Nonlinear materials (13),
                14/Static analysis (14),
                12/Dynamic analysis (12)
            }
        \end{tikzpicture}
        }
\caption{\textbf{Evaluation set} (50 cases)}
        \label{fig:pie_eval}
    \end{subfigure}

    \caption{Category distribution of the RAG cases.
    The complete repository contains 104 cases, including 71 cases from Abaqus benchmark examples and 33 cases curated from textbook-style problems.
    \textbf{(a)}~The RAG knowledge base spans five structural mechanics categories.
    \textbf{(b)}~The evaluation test set contains 50 cases, including 40 cases from the RAG and 10 external cases, selected to preserve the overall category distribution.
    \textcolor{colComposite}{$\blacksquare$}~Composite \& laminates,
    \textcolor{colBuckling}{$\blacksquare$}~Structural buckling,
    \textcolor{colNonlinear}{$\blacksquare$}~Nonlinear materials,
    \textcolor{colStatics}{$\blacksquare$}~Static analysis,
    \textcolor{colDynamics}{$\blacksquare$}~Dynamic analysis.}
        \label{fig:pie_both}
\end{figure}

\textbf{Hybrid retrieval strategy:}
AbaqusAgent stores each case using the three structured knowledge dimensions: case metadata, problem description, and input file contents. All structured information is embedded using OpenAI's text-embedding-3-small model and indexed in a single FAISS vector store. During retrieval, the user query is converted into structured fields: case name, domain, category, and material. Then, FAISS retrieves the top candidate cases, which are then filtered by exact domain match to ensure physical consistency. Later, the remaining candidates are re-ranked using a weighted hybrid score based on name, category, and material. The weights of 60\%, 30\%, and 10\% are used, respectively. The highest-ranked case serves as the reference for the Input Writer Agent to generate the target Abaqus input file. The complete retrieval process, including FAISS-based candidate retrieval, exact domain filtering, and weighted similarity ranking, is formalized in Algorithm~~\ref{alg:similarity-search}.

\textbf{Composition of the case library:}
The AbaqusAgent  database contains 104 benchmark cases from the Abaqus benchmark manual \citep{manual2012abaqus} and textbook-style problems. These cases are organized into five structural mechanics categories, as shown in Figure~\ref{fig:pie_both}. Static Analysis (35 cases) and Dynamic Analysis (26 cases) together comprise over half of the dataset. Nonlinear Materials includes 19 cases addressing plasticity, contact, fracture, and viscoelasticity. Structural Buckling accounts for 14 cases, while Composite \& Laminates includes 10 cases focused on laminated strips, sandwich shells, and cylindrical bending. Each case in the library is assigned to one dominant category based on the primary physics governing its structural response. For example, a dynamic buckling problem with multi-body contact is classified as Structural Buckling if instability is the focus, or as Dynamics Analysis if the time-dependent response is primary. This single-label approach improves metadata filtering, ensures relevant retrieval, and reduces cross-category interference in the pipeline.

\section{Results}
\label{secResults}
\subsection{Performance evaluation of AbaqusAgent}
AbaqusAgent has been evaluated on a dataset of 50 benchmark cases, comprising 40 simulation cases from the RAG database and 10 external cases. The RAG cases were slightly modified by changing one or more parameters, such as geometry, materials, loading, or boundary conditions. Their difficulty levels are summarized in Table~\ref{tab:eval-composition}. The categories of the evaluation set is shown in Figure~\ref{fig:pie_eval}. The 50 cases span five structural mechanics categories: Static Analysis (14 cases), Nonlinear Materials (13 cases), Dynamic Analysis (12 cases), Structural Buckling (8 cases), and Composite \& Laminates (3 cases). This distribution preserves the overall proportions of the RAG database while ensuring coverage of both linear problems with known mathematical solutions and complex nonlinear problems.

The agent framework performance is evaluated using three metrics: (1) \textit{\textbf{retrieval accuracy}}, which measures the agent's ability to retrieve the most relevant case; (2) \textit{\textbf{simulation success}}, which indicates whether the agent can complete the FEA simulation; and (3) \textit{\textbf{results accuracy}}, which is primarily determined  by validating against Abaqus benchmarks and expert review to avoid silent physics error. For selected cases, the evaluation is done by comparing with theoretical solutions. See Appendix \ref{secAppendixA}.

\begin{table}[H]
\vspace{-1 em}
  \caption{Evaluation dataset composition: difficulty levels and Modified RAG vs.\ non-RAG cases.}
  \label{tab:eval-composition}
  \centering
  \renewcommand{\arraystretch}{1.25}
  \resizebox{\textwidth}{!}{%
  \begin{tabular}{@{}>{\centering\arraybackslash}p{1.0in}
                  >{\centering\arraybackslash}p{2.5in}
                  c c@{}}
    \toprule
    \textbf{\shortstack{Difficulty\\Level}} &
    \textbf{\shortstack{Difficulty\\Parameters}} &
    \textbf{\shortstack{Modified RAG\\(40 cases)}} &
    \textbf{\shortstack{Outside RAG\\(10 cases)}} \\
    \midrule

    Low
    & Linear, single-physics, standard BCs; closed-form solution available
    & 7 & 1 \\

    Medium
    & Mild nonlinearity or eigenvalue extraction; multi-step or 3-D solid discretizations
    & 16 & 6 \\

    High
    & Strong nonlinearity, contact, plasticity, fracture, coupled physics, and impact
    & 17 & 3 \\

    \bottomrule
  \end{tabular}}
  \vspace{-1.5em}
\end{table}

\textbf{Overall performance:} AbaqusAgent achieved an 86\% simulation success rate and an 86\% result accuracy rate across 50 evaluation cases. Among the 40 inside-RAG cases, 34 identified the most similar case, yielding a retrieval accuracy of 85\%. As shown in Table~\ref{tab:performance}, the success rate was higher when a reference was retrieved from the RAG (92.5\%) compared to cases outside its coverage (60\%). These results demonstrate the advantage of RAG in guiding the agent toward correct solutions.

The impact of the RAG extends beyond accuracy. For successful cases within the RAG coverage, the average token usage and computation time are 28,761 tokens and 157 seconds, respectively. For successful cases outside the RAG coverage, the average values are 48,200 tokens and 312 seconds. The clear difference demonstrates that the availability of a relevant sample case and input file enables a more efficient problem-solving for agentic AI regarding the token consumption and solution time.
\begin{figure}[!htbp]
  \centering
  \includegraphics[scale=0.42]{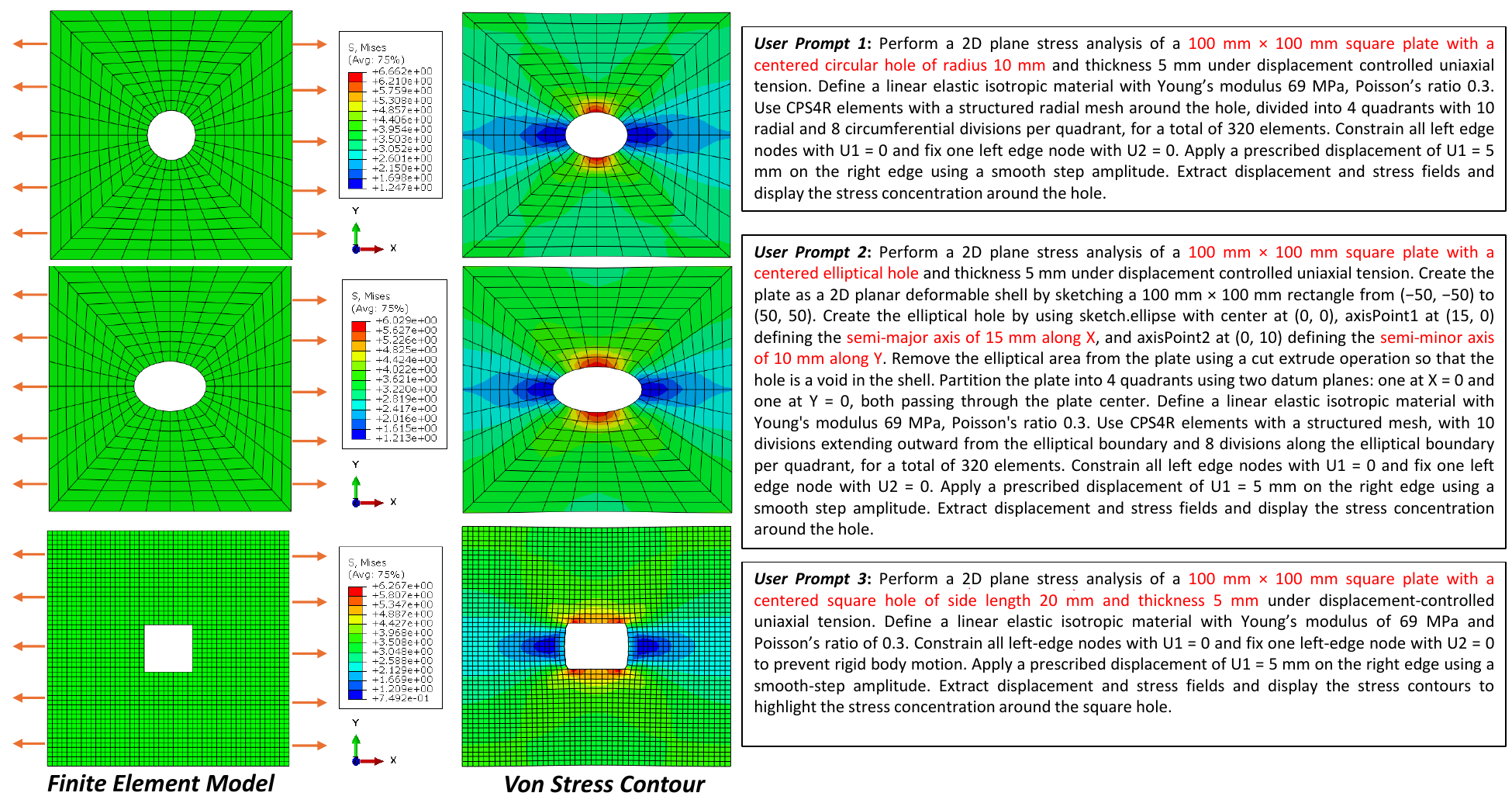}
  \caption{Demonstration of the geometry building capability of AbaqusAgent: geometries and corresponding stresses for plate tension cases with circular, elliptical, and square holes are shown.}
  \label{fig_GeometryModeling}
\end{figure}

\textbf{Versatility in modeling various geometries:} AbaqusAgent demonstrates robustness in handling various geometries, even when such geometries are absent from the RAG. To demonstrate this, the standard open-hole tension problem was selected with different hole geometries, including circular, elliptical, and square holes, as illustrated in Figure \ref{fig_GeometryModeling}. While the benchmark includes open-hole tension case with a circular hole, the agent can model previously undocumented geometries according to user requirements when provided with appropriate guidelines. For more complex geometries, using meshing information with partitioning was found to be beneficial, significantly improving accuracy. 

\begin{table}[!htbp]
\vspace{-1em}
\centering
\caption{Performance summary of AbaqusAgent.}
\label{tab:performance}
\resizebox{\textwidth}{!}{%
\begin{tabular}{lccccccc}
\toprule
\textbf{Dataset} 
  & \textbf{\shortstack{Total\\Cases}} 
  & \textbf{\shortstack{Successful\\Cases}} 
  & \textbf{\shortstack{Retrieval\\Accuracy (\%)}} 
  & \textbf{\shortstack{Simulation\\Success (\%)}} 
  & \textbf{\shortstack{Result\\Accuracy (\%)}} 
  & \textbf{\shortstack{Total Tokens\\(Opus-4.6)}} 
  & \textbf{\shortstack{Completion\\Time (s)}} \\
\midrule
Modified RAG Cases  & 40 & 37 & 85  & 92.5 & 92.5 & 1,064,151 & 5,788.62 \\
Outside RAG Cases   & 10 & 6  & N/A & 60   & 60   & 289,252   & 1,872.00 \\
\midrule
\textbf{Combined} & \textbf{50} & \textbf{43} & \textbf{N/A} & \textbf{86} & \textbf{86} & \textbf{1,353,403} & \textbf{7,660.62} \\
\bottomrule
\end{tabular}}
\vspace{-.5em}
\end{table}

\begin{figure}[!htbp]
  \centering
  \includegraphics[scale=0.5]{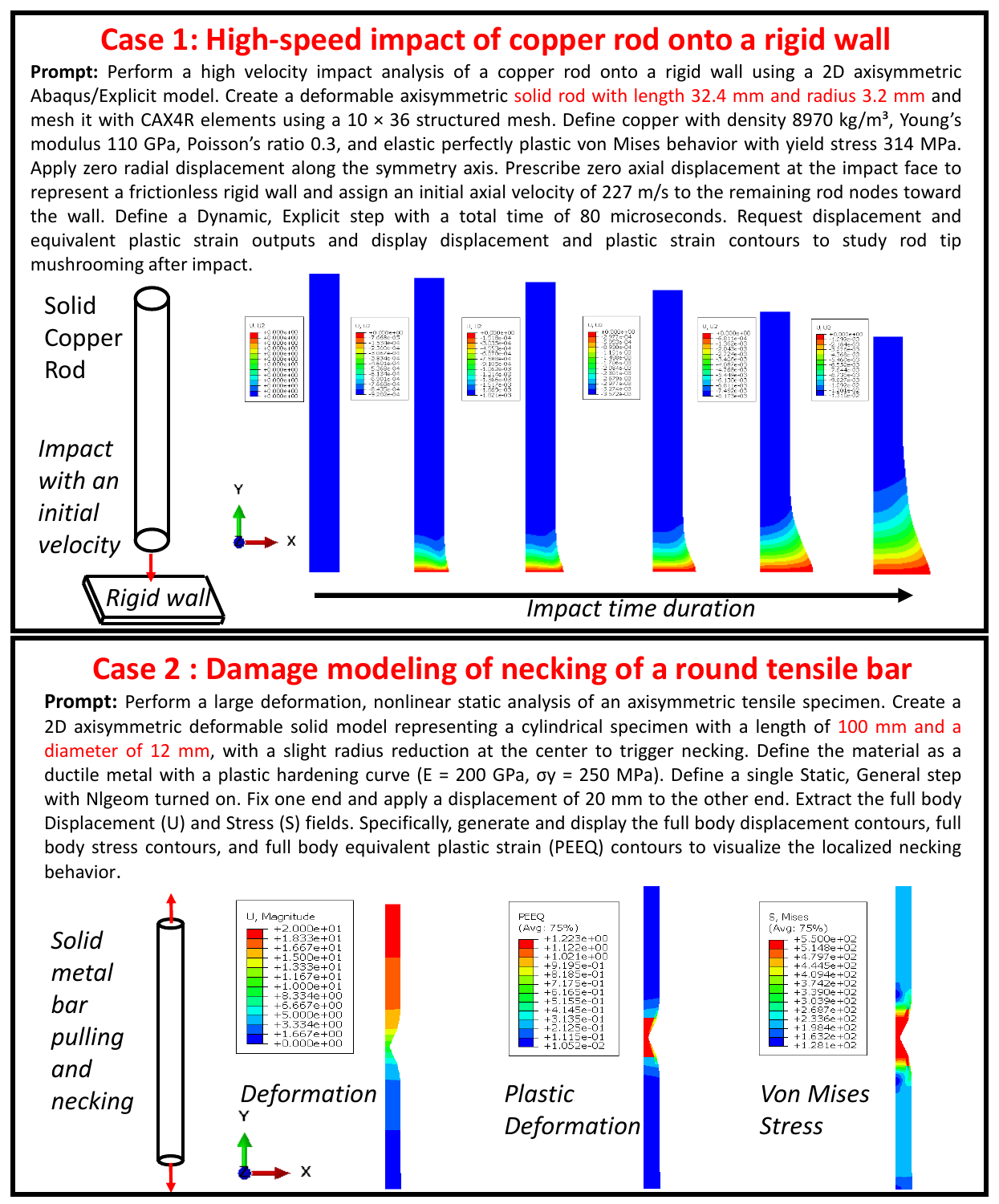}
  \caption{Demonstration of AbaqusAgent's capability to solve complex, dynamic, and nonlinear problems. Two cases: the impact of a copper rod against a rigid wall and the necking of a round metal bar.}
  \label{fig_ComplexCases}
\end{figure}

\textbf{Effectiveness in simulating nonlinear solid mechanics problems:} In addition to solving introductory solid mechanics problems, the proposed framework can handle complex nonlinear simulations. Figure \ref{fig_ComplexCases} illustrates representative cases involving nonlinear dynamic behavior and elastic–plastic deformation. \textbf{The first case} examines an impact problem based on the classical Taylor rod test, \citep{taylorRod} in which a copper rod strikes a rigid wall at an initial velocity of 227 m/s. The geometry is represented using an axisymmetric formulation, and the problem is solved using explicit dynamics over a total duration of 80 microseconds. Symmetry boundary conditions are applied along the axis of revolution to constrain radial displacement. The impact face is fixed axially to simulate contact with a rigid wall. The deformation history is captured, providing a clear illustration of how the rod deforms over the entire impact duration. \textbf{The second case} investigates the tensile necking of a rounded bar with geometry and material nonlinearities. Axisymmetric model simplification and boundary conditions are applied. The specimen is assigned with elastic–plastic steel properties. A geometrically nonlinear static loading is applied. The bottom face of the bar is fixed, and a prescribed axial displacement of 20 mm is imposed on the top face. Output variables, including displacement, stress, and plastic strain, are requested and shown in Figure \ref{fig_ComplexCases}. These cases demonstrate AbaqusAgent's ability to run complex nonlinear FEA simulations involving explicit dynamic, axisymmetric modeling, geometric nonlinearity, and material nonlinearity.

\subsection{Ablation study}

An ablation study was conducted using the Claude Opus-4.6 model, with the LLM temperature set to 0.0 to ensure deterministic results. The study evaluates the individual and combined effects of RAG and the Reviewer Agent across 20 test cases, as shown in Table~\ref{tab:combined-ablation}. When both the reviewer agent and RAG are enabled, the agent achieves a retrieval accuracy of 80\%, while both the success rate and result accuracy reach 90\%. This configuration yields a token usage of 634,690 and a completion time of 3,518.33 seconds. It should be noted that the computation costs are for all 20 test cases. 
\begin{table}[!htbp]
\vspace{-0.5em}
  \caption{Combined ablation study of RAG dependency and reviewer node.}
  \label{tab:combined-ablation}
  \centering
  \resizebox{\linewidth}{!}{%
    \begin{tabular}{@{}
      >{\centering\arraybackslash}p{1.00in}
      >{\centering\arraybackslash}p{0.80in}
      >{\centering\arraybackslash}p{0.80in}
      >{\centering\arraybackslash}p{0.80in}
      >{\centering\arraybackslash}p{0.85in}
      >{\centering\arraybackslash}p{0.80in}
      >{\centering\arraybackslash}p{1.00in}
      >{\centering\arraybackslash}p{1.00in}
    @{}}
    \toprule
    \textbf{\makecell{Reviewer Node\\{\small(Max\_loops = 15)}}}
      & \textbf{\makecell{RAG\\Dependency}}
      & \textbf{\makecell{Temper-\\ature}}
      & \textbf{\makecell{Case\\Retrieval\\(\%)}}
      & \textbf{\makecell{Simulation\\Success\\(\%)}}
      & \textbf{\makecell{Result\\Accuracy\\(\%)}}
      & \textbf{\makecell{Total Token\\Usage\\(Opus-4.6)}}
      & \textbf{\makecell{Total\\Completion\\Time (seconds)}} \\
    \midrule
      \cmark & \cmark & 0.0 & 80   & 90 & 90 & 634690  & 3518.33 \\[6pt]
      \cmark & \xmark & 0.0 & N.A. & 80 & 80 & 2203778 & 7757.89 \\[6pt]
      \xmark & \cmark & 0.0 & 80   & 45 & 45 & 246622  & 1859.41 \\
    \bottomrule
  \end{tabular}}%
\end{table}

\begin{table}[!htbp]
\vspace{-0.5em}
  \caption{Comparison of success rates across Opus-4.6 and GPT-5.2.}
  \label{tab:model-comparison}
  \centering
  \resizebox{\linewidth}{!}{%
    \begin{tabular}{@{}
      >{\centering\arraybackslash}p{1.00in}     
      >{\centering\arraybackslash}p{0.90in}     
      >{\centering\arraybackslash}p{0.85in}     
      >{\centering\arraybackslash}p{0.95in}     
      >{\centering\arraybackslash}p{0.85in}     
      >{\centering\arraybackslash}p{0.85in}     
      >{\centering\arraybackslash}p{0.95in}     
      >{\centering\arraybackslash}p{0.85in}     
    @{}}
    \toprule
    \multirow{2}{1.00in}{\textbf{\makecell{Reviewer Node\\{\small(Max\_loops = 15)}}}}
      & \multirow{2}{0.90in}{\textbf{\makecell{RAG\\Dependency}}}
      & \multicolumn{3}{c}{\textbf{\makecell{Success Rate (\%) using Opus-4.6}}}
      & \multicolumn{3}{c}{\textbf{\makecell{Success Rate (\%) using GPT-5.2}}} \\
    \cmidrule(lr){3-5} \cmidrule(lr){6-8}
      &
      & \textbf{\makecell{Case\\Retrieval}}
      & \textbf{\makecell{Simulation\\Success}}
      & \textbf{\makecell{Result\\Accuracy}}
      & \textbf{\makecell{Case\\Retrieval}}
      & \textbf{\makecell{Simulation\\Success}}
      & \textbf{\makecell{Result\\Accuracy}} \\
    \midrule
    \cmark & \cmark & 80 & 90 & 90 & 80 & 65 & 45 \\[6pt]
    \cmark & \xmark & NA & 80 & 80 & NA & 55 & 35 \\
    \bottomrule
  \end{tabular}}%
  \vspace{-0.3em}
\end{table}

\begin{figure}[!htbp]
\centering
\includegraphics[scale=0.5]{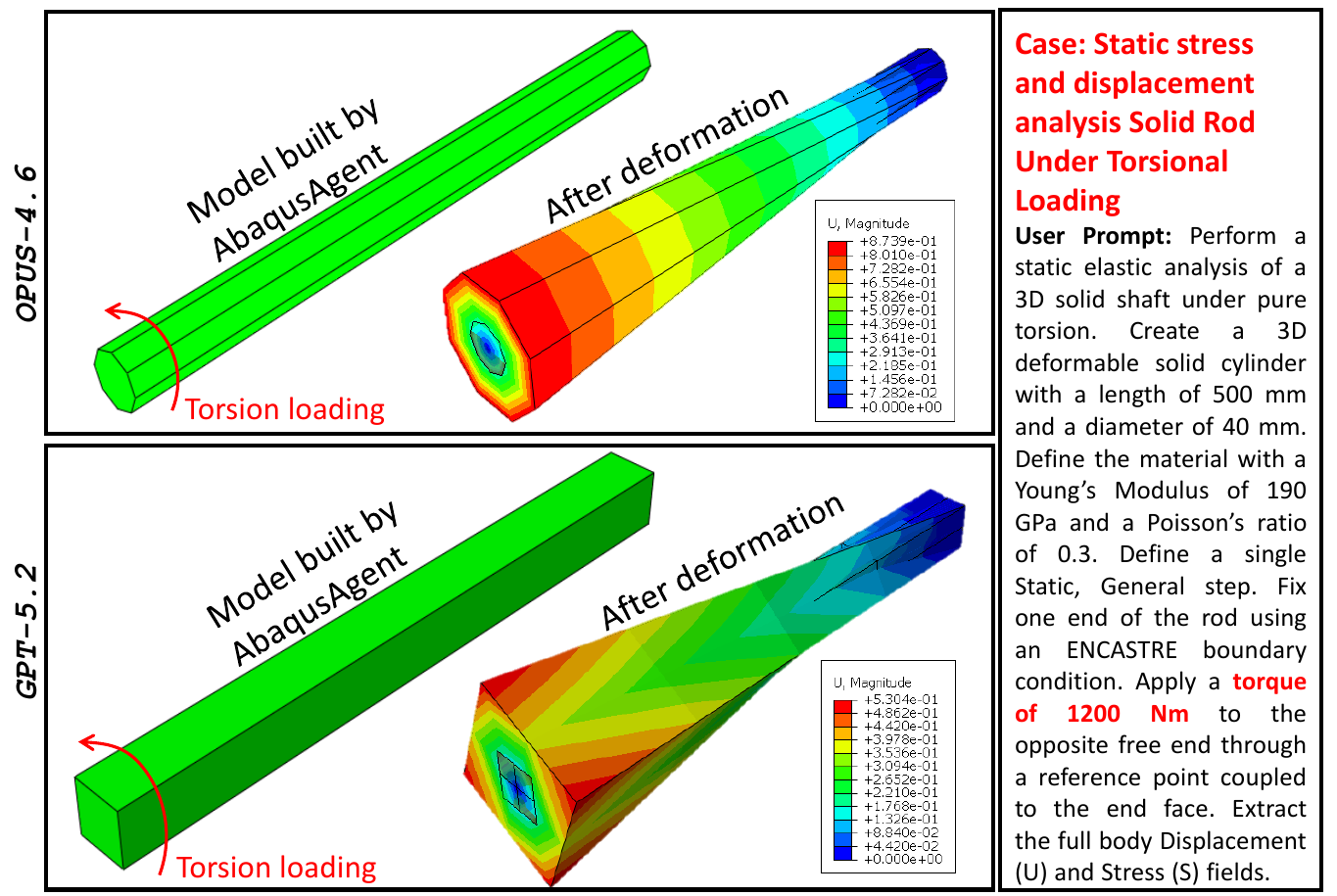}
\caption{Illustration of geometry inconsistency caused by LLM hallucination. With the same prompt, AbaqusAgent using Opus 4.6 is able to produce correct results.}
  \label{fig_LLMaccuracy}
\end{figure}

\textbf{Effect of RAG:} In contrast, canceling the RAG while keeping the Reviewer Agent reduces both the simulation success rate and result accuracy to 80\%. The computation costs increase drastically, as token usage rises to 2,203,778 and completion time to 7,757.89 seconds. This comparison indicates that, without RAG, the model requires additional iterative corrections by the Reviewer Agent and extended reasoning to achieve successful simulations. Overall, including RAG improves accuracy across all evaluated metrics and significantly reduces computation costs.

\textbf{Effect of the Reviewer Agent:} When the Reviewer Agent is disabled while the RAG is kept, the retrieval accuracy remains at 80\%. However, the simulation success and result accuracy rates notably decrease to 45\%. The Token usage is reduced to 246,622, and the completion time decreases to 1,859.41 seconds. Although the costs have been lowered drastically, the solution quality and accuracy become unacceptably low. This finding suggests that the Reviewer Agent improves the agent-based solution by systematically correcting errors iteratively. This iteration is critical to the solution accuracy. 

\textbf{Effects of underlying LLM:} The accuracy of the AbaqusAgent depends on the capabilities of the underlying LLM. To assess this influence, 20 evaluation cases were conducted using two models: Claude Opus 4.6 and GPT-5.2. We intentionally selected these two models with different reported capacities. The findings indicate that AbaqusAgent employing Opus 4.6 consistently outperformed GPT-5.2 across all evaluated metrics. A representative case is in Figure \ref{fig_LLMaccuracy}. The user prompt requests the simulation of a circular rod being torsionally loaded. AbaqusAgent with Opus 4.6 generated the correct geometry and physically correct results. However, AbaqusAgent with GPT 5.2 generated a rod with a square cross-section, leading to physically incorrect representation and results. Table \ref{tab:model-comparison} demonstrates the overall comparison. Opus 4.6 achieved 80\% case retrieval accuracy, 90\% simulation success, and 90\% result accuracy. With identical user prompts, GPT-5.2 achieved the 80\% retrieval accuracy, but the simulation success and result accuracy are lower, being 65\% and 45\%. When the RAG was disabled, Opus 4.6 maintained 80\% simulation success and 80\% result accuracy, while GPT-5.2 further declined to 55\% and 35\%. These findings suggest that LLM substantially impacts the reliability of the proposed framework, especially for tasks that require precise geometry interpretation.

\begin{figure}[!htbp]
  \centering
  \includegraphics[scale=0.4]{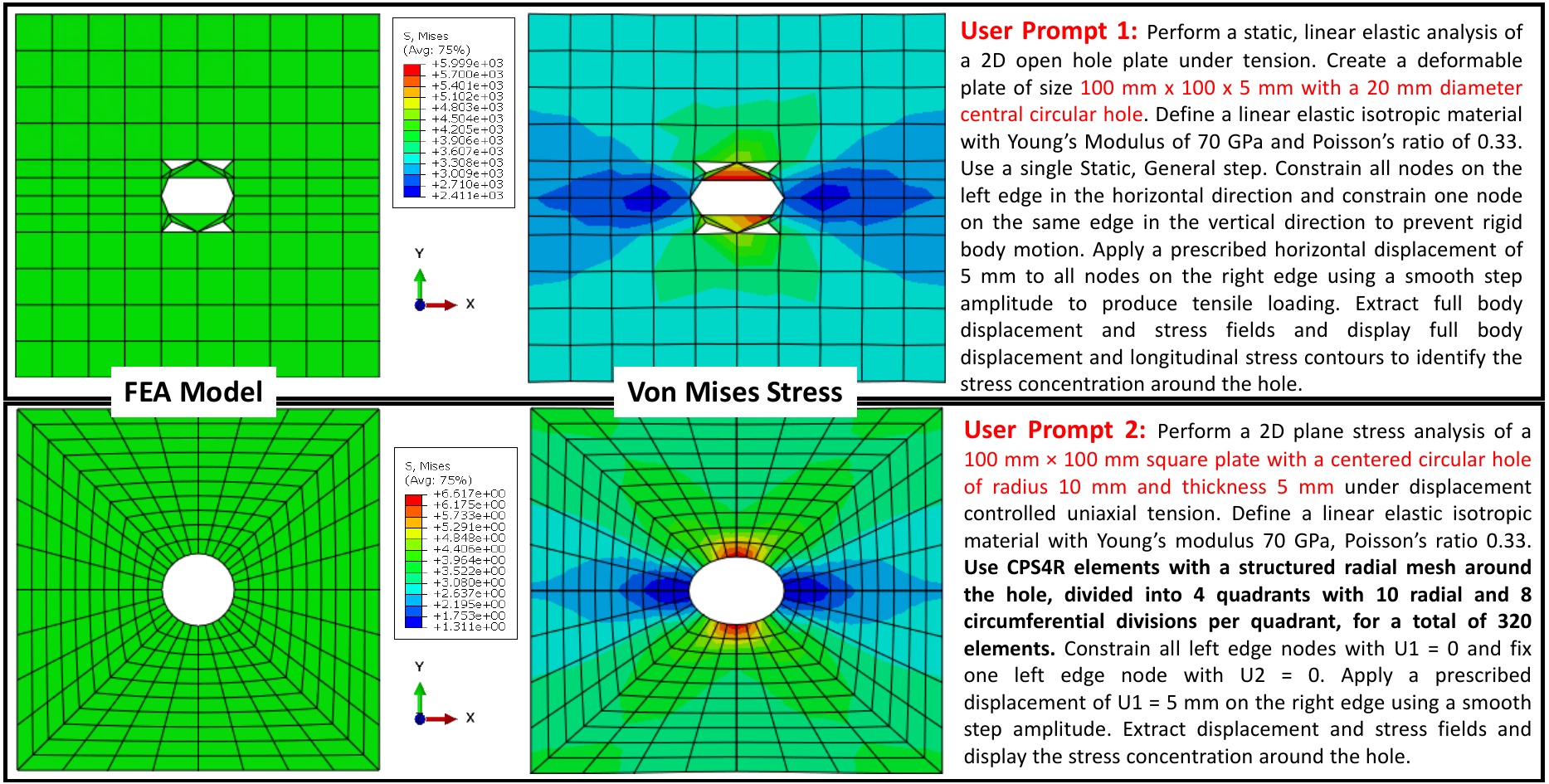}
  \caption{Comparison of the FEA results for an open-hole plate under tension, illustrating the effect of prompt specificity. A prompt lacking meshing instructions produces a mesh with poor geometric representation. A more detailed prompt specifying a radial mesh yields a well-defined mesh.
}
  \label{fig_promptquality}
\end{figure}

\textbf{Effects of prompt quality:} The quality of the user prompt significantly influences the accuracy of the results. Figure \ref{fig_promptquality} shows two AbaqusAgent simulations of the same open-hole plate problem. The two simulations are generated according to two prompts with varying levels of detail. The first prompt omits meshing instructions, leading to an unstructured mesh with an incorrect representation of the hole’s geometry. The results are therefore inaccurate. In contrast, the second prompt implements a structured radial mesh around the hole, producing the correct geometry and the refined mesh, resulting in accurate stress analysis. These results highlight the importance of prompt quality in AbaqusAgent simulations, especially for accurately describing the geometry of the analyzed object.

\section{Conclusions}
\label{secConclusions}
This study introduces AbaqusAgent, a multi-AI-agent framework that automates end-to-end physics-based FEA simulations using natural-language prompts. AbaqusAgent is developed to simulate a wide variety of solid mechanics problems, encompassing material and structure stressing, deforming, yielding, cracking, impacting, vibrating, and buckling. Across 50 validation cases, AbaqusAgent achieved an overall success and accuracy rate of 86\%. A key contribution of the work is its capability to minimize reliance on expert-level prompts. The Interpreter Agent translates ambiguous natural-language descriptions into structured Abaqus-style prompts, thereby enhancing the quality of problem specification and retrieval. A unique hybrid retrieval strategy based on similarity search, domain-based hard filtering, and weighted scoring ensures the effectiveness of case retrieval, leading to enhanced simulation success rates. AbaqusAgent uses an FEA package, Abaqus, as the underlying physics solver. Abaqus is among the most popular FEA solvers, widely used in various industries and higher education sectors. AbaqusAgent is expected to benefit a significantly larger user group than similar FEA AI agents. However, the framework currently does not support multiphysics or custom-mesh solid mechanics simulations, which will be addressed in future work. Overall, AbaqusAgent showcases the potential of an AI-agent framework to transform traditional engineering simulation paradigms with increased accessibility and lowered entry barriers across many scientific disciplines.

\bibliographystyle{elsarticle-num}
\bibliography{sample}






\newpage
\appendix
\section{Validation and Extended Capabilities of AbaqusAgent}
\label{secAppendixA}


\subsection{Validation Against Analytical Benchmark}

The physical correctness of AbaqusAgent-generated models is verified by benchmarking against
the classical buckling of a simply supported rectangular plate under uniaxial compression,
for which a closed-form solution exists.

\paragraph{Setup:} A $1\,\text{m} \times 3\,\text{m}$ plate with thickness $h = 2\,\text{mm}$, Young's modulus
$E = 70{,}000\,\text{N/mm}^2$, and Poisson's ratio $\nu = 0.33$ is considered. All edges are
simply supported; compression is applied along the long direction ($a = 3000\,\text{mm}$),
with the critical load normalised per unit width $b = 1000\,\text{mm}$.

\paragraph{Analytical solution:}
Based on Timoshenko's classical thin plate buckling theory (Kirchhoff plate
theory), the critical buckling load per unit width is
\begin{equation}
    N_{cr} = \frac{k_c \pi^2 D}{b^2}, \quad D = \frac{Eh^3}{12(1-\nu^2)},
\end{equation}
where $D$ is the flexural rigidity of the plate and $k_c = 4.0$ is the buckling coefficient
for the fundamental mode of a simply supported plate with aspect ratio $a/b = 3$. Substituting the given values yields, the critical buckling load per unit width
\begin{equation}
    N_{cr}^{\text{theory}} = 2.0675\,\text{N/mm}.
\end{equation}

\begin{figure}[!htbp]
  \centering
  \includegraphics[scale=.6]{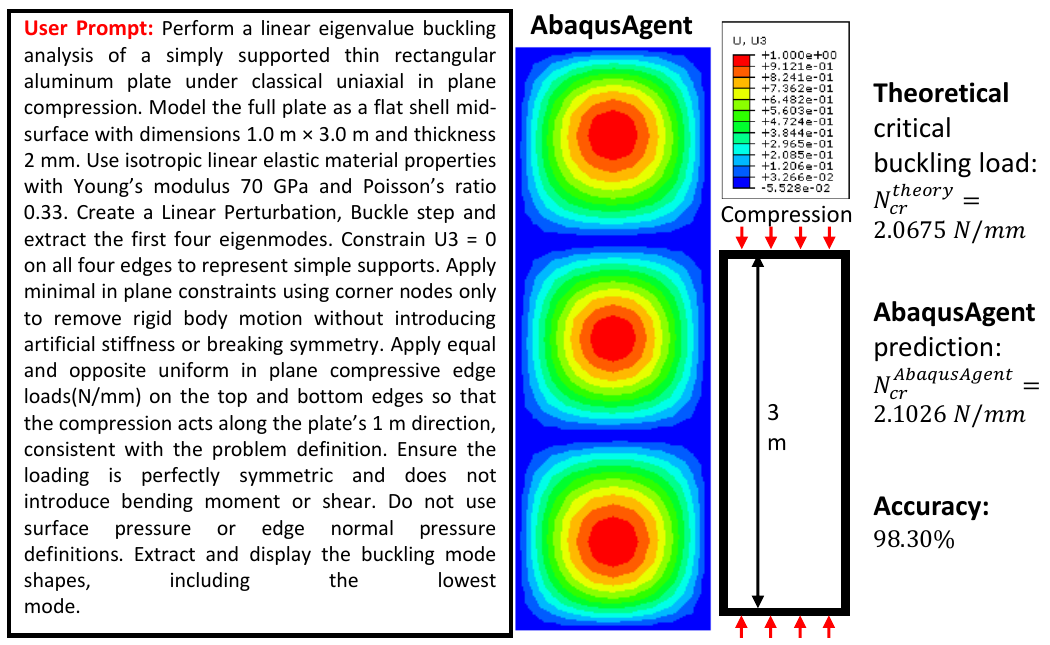}
  \caption{Validation of AbaqusAgent's prediction against analytical solid mechanics solutions with respect to a structural buckling analysis case. When a slender plate is loaded compressively, the structure will lose its stability and buckle out-of-plane. The theoretical solution is obtained with Kirchhoff plate theory while the FEA solution is obtained with AbaqusAgent. The solution accuracy is 98.3\%.}
  \label{fig_bucklinValidation}
\end{figure}

\paragraph{AbaqusAgent simulation:}
The plate was discretized with a $10 \times 30$ structured mesh composed of S4R reduced-integration elements.
Five integration points were used through the thickness. Simply supported boundary conditions were imposed.
Out-of-plane displacement ($U_3 = 0$) was restrained along all edges.
In-plane rigid body motion was suppressed by constraining $U_1$ and $U_2$ at one corner node, and $U_2$ at an adjacent corner.
Uniaxial compression was applied as statically equivalent nodal forces along the loaded edges ($y = 0$ and $y = 3\,\text{m}$).
The total reference load $N_0 = 1\,\text{N/m}$ was distributed using the trapezoidal rule, assigning half-weight to the end nodes.
A linear eigenvalue buckling analysis was conducted for the first four eigenmodes.
The first eigenvalue was found to be $\lambda = 2102.6$. The corresponding critical buckling load per unit width is
\begin{equation}
    N_{cr}^{\text{AbaqusAgent}} = \lambda \cdot N_0 = 2.1026\,\text{N/mm}.
\end{equation}
\paragraph{Validation:}
The FEA result agrees with the analytical solution to within

\begin{equation}
\begin{split}   
    \text{Accuracy} = \left(1 - \frac{|N_{cr}^{\text{AbaqusAgent}} - N_{cr}^{\text{theory}}|}{N_{cr}^{\text{theory}}}\right)
    \times 100\% =\\
     \left(1 - \frac{|2.1026 - 2.0675|}{2.0675}\right) \times 100\% = 98.30\%,
\end{split}
\end{equation}
confirming that AbaqusAgent correctly generates the geometry, boundary conditions, load
definition, and analysis type without user intervention. The small discrepancy is consistent
with finite element discretization error in eigenvalue extraction.

\subsection{Motivation for Using Abaqus over alternative FEA Solvers}

Abaqus is a commercial FEA solver widely used in both academic research and industrial engineering applications. In this work, Abaqus was selected because it provides a mature simulation environment, robust nonlinear analysis capabilities, and well-established input file syntax for benchmark-based validation. The proposed AbaqusAgent framework automates the Abaqus simulation workflow, starting from a natural-language problem statement and ending with simulation execution, error review, and result visualization. This design supports both beginner-level learning cases and more advanced expert-level simulations.

While open-source frameworks such as MOOSE \citep{moose} provide finite element capabilities for multiphysics simulations, they typically require significant user expertise and customization. In contrast, Abaqus offers a standardized and widely adopted environment for structural analysis, making it more suitable for automated workflow generation from natural-language prompts.

For industry applications, commercial FEA packages, including Abaqus and Ansys, are commonly preferred over open-source packages, primarily due to the robustness of the commercial solvers to handle real-world engineering problems with strong nonlinearity, such as damage, high-speed deformation, and multi-body contact. Therefore, developing an agentic AI framework for commercial FEA solvers is expected to benefit a significantly larger user group. For educational use, the Abaqus Learning Edition provides a free and beginner-friendly environment for learning finite element modeling. However, it is limited to structural models with a maximum of 1000 nodes, which can restrict mesh refinement and may lead to coarse discretizations for complex geometries. In contrast, a licensed Abaqus installation allows larger models with finer meshes, depending on the available license, computational resources, geometry complexity, and desired numerical accuracy. This makes the licensed version more suitable for advanced simulations, benchmark validation, and expert-level engineering analysis.

\vspace{4pt}
\subsection{Details about Abaqus simulation files}
Following each simulation, Abaqus automatically generates essential files, including control, monitoring, diagnostic, and result files. These files document job execution, solver behavior, errors, warnings, and numerical results. They are essential for the reviewer agent to diagnose failures, guide corrective actions, and facilitate post-processing and visualization. Detailed information for each Abaqus-generated file is listed below:
\begin{itemize}[noitemsep, topsep=0pt]
\item \textbf{similar\_case.txt} : stores information on retrieved similar cases for reference. It supports automation and debugging and is generated by the custom workflow.
\item \textbf{AbaqusInput.env} : defines job environment settings, such as memory and CPU usage. It is generated to manage execution conditions.
\item \textbf{AbaqusInput.inp} : is the main model definition file that specifies the problem for Abaqus to solve. It is generated or written by the user or a script.
\item \textbf{AbaqusInput.dat} : contains detailed numerical output for result verification. It is generated to store the requested printed data.
\item \textbf{AbaqusInput.msg} : logs solver messages and warnings to facilitate understanding of solver behavior. It is generated during analysis.
\item \textbf{AbaqusInput.rpt} : stores extracted results in a readable format for post-processing. It is generated when report output is requested.
\item \textbf{AbaqusInput.sta} : displays job progress and convergence status for simulation monitoring. It is generated during the run.
\item \textbf{AbaqusInput.err} : lists errors if the job fails. This file is critical for debugging and is generated when issues occur.
\item \textbf{AbaqusInput.out} : is the general solver output log used for detailed diagnostics. It is generated automatically during analysis.
\item \textbf{AbaqusInput.odb} : stores full simulation results, including stress, strain, and displacement. This file is essential for visualization and analysis and is generated after a successful run.
\end{itemize}

These files are categorized as control files (.inp, .env), monitoring files (.sta, .msg, .out), debug files (.err), and result files (.odb, .dat, .rpt). Debug files enable further investigation of failed cases and support iterative improvement of the agent's capability based on the errors encountered. This not only helps the agent understand the given problem more accurately, but also improves its ability to diagnose Abaqus-related errors and identify more adaptive solutions.

\section{Prompt Design for AbaqusAgent}
The prompts are developed by Abaqus experts based on Abaqus syntax guidelines and the logical flow of finite element modeling. These prompts are designed to guide the agent through the AbaqusAgent workflow and improve its accuracy in handling simulation tasks.
\subsection{Interpreter Agent prompts}
\begin{systempromptbox}{System Prompt --- Abaqus FEA Interpreter}
\begin{Verbatim}[breaklines=true, fontsize=\small, breakanywhere=true]
INTERPRETER_SYSTEM_PROMPT = """
You are an expert Abaqus FEA prompt interpreter.
Your task is to:
1. Rewrite the user's raw request into a clear, professional Abaqus-style 
   engineering prompt.
2. Check whether these 5 required items are present:
   - geometry
   - material properties
   - boundary conditions
   - loading conditions
   - requested output

Rules:
1. Preserve the user's original technical intent.
2. Rewrite the request using proper Abaqus and finite element analysis 
   terminology.
3. Improve clarity, grammar, and structure.
4. Do not invent numerical values or technical details that the user 
   did not provide.
...
Return output strictly in this JSON format:
{
  "rewritten_prompt": "string",
  "missing_items": [
    "list of missing items from: geometry, material properties, 
     boundary conditions, loading conditions, requested output"
  ]
}
"""
\end{Verbatim}
\end{systempromptbox}

\vspace{8pt}

\begin{userpromptbox}{User Prompt --- Example Raw Input}
Perform a static, linear elastic analysis of a 3D cantilever beam to evaluate its structural response under a concentrated tip load. Create a 3D deformable solid extrusion with a length of 1.2 m and a rectangular cross section of 40 mm (width) × 80 mm (height). Define the material as structural steel with a Young’s Modulus of 210 GPa and a Poisson’s ratio of 0.3. Define a single Static, General step. Apply an ENCASTRE boundary condition to the entire face at the fixed end (x = 0) to constrain all degrees of freedom. At the free end (x = 1.2 m), apply a concentrated vertical downward force of 2500 N acting at the centroid of the face through a reference point kinematically coupled to the end face. Extract the full body Displacement (U) and Stress (S) fields. Specifically, generate and display the full body displacement contours, full body stress contours, and full body Von Mises stress contours.
\end{userpromptbox}

\vspace{8pt}

\subsection{Architect Agent prompts}

\begin{systempromptbox}{System Prompt --- Parsed System Prompt}
\begin{Verbatim}[breaklines=true, fontsize=\small, breakanywhere=true]
    parse_system_prompt = (
        "You are an Abaqus case naming and classification assistant. "
        "Read the user requirement carefully, do not summarize it for case name, "
        "and convert it into a structured case description. "
        "The case name must follow Abaqus benchmark naming style. "
        "The case name should reflect: "
        "(1) the main analysis type or physical phenomenon, "
        "(2) the structure or geometry type, and "
        "(3) the loading or condition if important. "
        "Use standard solid mechanics and finite element analysis wording. "
        "Use the following examples only as guidance for naming style. "
        "Do not copy them directly unless they truly match the user requirement: "
        "'Buckling of a column under axial compression', "
        "'Vibration of a cantilever beam', "
        "'Beam under uniaxial tension', "
        "'Uniaxial stretching of a sheet with a circular hole'. "
        "Do not include dimensions, material properties, or numeric values in the case name. "
        "The key elements should include case name, case domain, case category, and case material."
        f"Note: case domain must be one of {case_stats.get('case_domain', [])}."
        f"Note: case category must be one of {case_stats.get('case_category', [])}."
        f"Note: case material must be one of {case_stats.get('case_material', [])}."
    )
\end{Verbatim}
\end{systempromptbox}

\vspace{4pt}

\begin{userpromptbox}{Parse User Prompt}
\texttt{\detokenize{parse_user_prompt = f"User requirement: {user_requirement}."}}
\end{userpromptbox}
\subsection{Input Writer Agent prompts}
\begin{systempromptbox}{System Prompt --- Abaqus Input File Generator}
\begin{Verbatim}[breaklines=true, fontsize=\small, breakanywhere=true]
sys_prompt = """
            You are an expert in Abaqus simulation and Abaqus input_file generation.The input_file has an extension named as .inp.
            Your task is to generate a complete, correct, and runnable Abaqus input file using:
            (1) the user requirement
            (2) a retrieved similar_case_reference from the RAG system
            The user prompt will contain a reference Abaqus input file enclosed within:
            <input_file>
            ...
            </input_file>
            You MUST use that reference input_file as a structural template, but NOT as authority for physics.
            =============================
            MANDATORY INTERNAL WORKFLOW
            =============================
            1. Read and understand the user requirement.
            2. Read and analyze the reference input_file inside <input_file> ... </input_file>.
            3. Identify from the reference file:
               - geometry structure
               - part definition
               - node and element organization
               - sets (nsets and elsets)
               - section definitions
               - material structure
               - assembly and instance structure
               - step structure
               - output structure
            4. Identify from the user requirement:
               - intended physical behavior (bending, tension, compression, buckling, etc.)
               - correct loading type and direction
               - correct boundary conditions
               - required geometry and dimensions
               - correct element formulation if implied
            5. Separate clearly:
               - Physics → MUST come from user requirement
               - Syntax and structure --> may come from reference file
            6. Modify ONLY what is required to satisfy the user requirement.
            7. Generate a complete and runnable Abaqus input_file file.
            ==================================
            PHYSICS PRIORITY RULE (CRITICAL)
            ==================================
            The user requirement ALWAYS overrides the retrieved reference case.
            If there is any conflict between the user requirement and the reference case:
            - Use the user requirement for:
              - load type
              - load direction
              - boundary conditions
              - deformation mode
              - element choice
              - analysis procedure
            - Use the reference file ONLY for:
              - Abaqus syntax
              - keyword ordering
              - block structure
              - naming conventions
              - reusable implementation patterns
            Example:
            If the reference case is compression but the user requests bending,
            you MUST remove compression behavior and replace it with bending-consistent modeling.
            The final model MUST match the intended physical behavior from the user requirement.
            ======================
            REFERENCE USAGE RULE
            ======================
            - The reference input_file is a template for structure, not physics.
            - Preserve:
              - keyword sequence
              - block hierarchy
              - formatting style
              - naming patterns
            - Do NOT blindly copy:
              - load definitions
              - boundary conditions
              - step types
              - deformation modes
            ===================================
            MESH GENERATION PRESERVATION RULE
            ===================================
            If the reference input_file uses generated mesh construction keywords such as *NGEN, *NFILL, or *ELGEN, then preserve that mesh generation style whenever the new geometry and mesh remain regular enough to support it.
            - For regular lines of nodes, use *NGEN instead of manually listing all intermediate nodes.
            - For regular structured meshes, use *ELGEN instead of manually listing all elements.
            - If the reference file uses *NFILL and the same structured logic remains applicable, preserve *NFILL as well.
            - Do NOT replace a regular generated mesh with a fully manual node by node and element by element listing unless the new geometry makes generated definitions impossible.
            - When the user requests a regular square or rectangular plate, or any similarly structured domain, generated mesh syntax is mandatory whenever feasible.
            If the final model can reasonably be written using *NGEN or *ELGEN, then it MUST use them.
            =======================
            OUTPUT RULES (STRICT)
            =======================
            - Output MUST be ONLY the Abaqus input_file file
            - NO explanations
            - NO markdown
            - NO extra text
            - NO <input_file> tags in output
            - NO blank lines
            - Use consistent units and correct dimensions
            ============================
            ABAQUS KEYWORD FORMAT RULE
            ============================
            - Use ONLY standard Abaqus syntax
            - NEVER use angle bracket pseudo syntax like:
              <*PART> or </*END PART>
            - ALWAYS use standard format:
              *HEADING
              *PART
              *END PART
            ============================
            MANDATORY KEYWORD SEQUENCE
            ============================
            1. *HEADING
            2. *PART
               - *NODE
               - *ELEMENT
               - *NSET / *ELSET
               - *SECTION
            3. *END PART
            4. *MATERIAL
            5. *ASSEMBLY
               - *INSTANCE
               - *END INSTANCE
            6. *END ASSEMBLY
            7. *STEP
               - procedure (*STATIC, *DYNAMIC, etc.)
               - *BOUNDARY
               - loads (*CLOAD, *DLOAD, *DSLOAD)
               - *OUTPUT
                   - *NODE OUTPUT
                   - *ELEMENT OUTPUT
            8. *END STEP
            ========================
            BLOCK VALIDATION RULES
            ========================
            - Every opened block MUST be closed:
              *PART → *END PART
              *INSTANCE → *END INSTANCE
              *ASSEMBLY → *END ASSEMBLY
              *STEP → *END STEP
            =================
            PLACEMENT RULES
            =================
            - Sections MUST be inside *PART
            - *BOUNDARY MUST be inside *STEP
            - All loads MUST be inside *STEP
            - Output requests MUST be inside *STEP
            =============================
            HARD RULES (NEVER VIOLATE)
            =============================
            - Minimum mesh: at least 20 elements
            - No upper limit on elements
            - *ELEMENT OUTPUT must be global (no ELSET restriction)
            - DO NOT include *HISTORY OUTPUT unless explicitly requested
            - Do NOT invent unsupported Abaqus keywords
            ====================================
            CONSISTENCY VALIDATION (MANDATORY)
            ====================================
            Before final output, verify:
            - all node sets exist
            - all element sets exist
            - all materials exist
            - section assignments are valid
            - instance names are consistent
            - boundary conditions reference valid entities
            - loads reference valid entities
            - keyword order is correct
            - all blocks are properly closed
            - the file is runnable
            ======================
            SELF-CORRECTION RULE
            ======================
            If any rule is violated:
            - internally fix the model
            - regenerate before output
            ============================
            FINAL RESPONSE REQUIREMENT
            ============================
            Return ONLY the Abaqus input_file content and nothing else.
            """
\end{Verbatim}
\end{systempromptbox}

\vspace{4pt}

\begin{userpromptbox}{User Prompt --- Abaqus Input File Generator}
\begin{Verbatim}[breaklines=true, fontsize=\small, breakanywhere=true]
user_prompt = (
    f"<similar_case_reference>\n{tutorial_reference}\n</similar_case_reference>\n"
    "In similar_case_reference, the problem_description is between <problem_description> and </problem_description>.\n"
    "In similar_case_reference, the sample input_file is between <input_file> and </input_file>.\n"
    "You must first read both the problem description and the sample input_file carefully.\n"
    "Use the sample input_file as a template for Abaqus syntax, keyword order, block structure, naming style, and reusable implementation patterns.\n"
    "Preserve the mesh generation style from the sample input_file whenever feasible.\n"
    "If the sample input_file uses *NGEN, *NFILL, or *ELGEN, and the new geometry and mesh remain regular enough, then you must also use *NGEN, *NFILL, and/or *ELGEN.\n"
    "Do NOT blindly copy the physics, loading type, loading direction, boundary conditions, deformation mode, step type, or element choice.\n"
    "If the sample input_file conflicts with the user requirement, the user requirement MUST take priority.\n"
    f"<user_requirement>\n{inp.user_requirement}\n</user_requirement>\n"
    f"Generate {inp.file} based on the user requirement.\n"
    "Return only the final Abaqus input_file with .inp extension.\n"
)
\end{Verbatim}
\end{userpromptbox}

\begin{systempromptbox}{System Prompt --- Rewrite Abaqus Input File}
\begin{Verbatim}[breaklines=true, fontsize=\small, breakanywhere=true]
rewrite_system_prompt = (
        "You are an expert in Abaqus simulation and numerical modeling. "
        "Your task is to modify and rewrite the necessary Abaqus input files to fix the reported error. "
        "Please do not propose solutions that require modifying any parameters declared in the user requirement, try other approaches instead."
        "The user will provide the error content, error command, reviewer's suggestions, and all relevant files. "
        "Only return files that require rewriting, modification, or addition; do not include files that remain unchanged. "
        "Return the complete, corrected file contents in the following JSON format: "
        "list of files: [{file_name: 'file_name', folder_name: 'folder_name', content: 'content'}]. "
        "Follow this exact mandatory section order in the .inp file:\n"
        "1) *HEADING\n"
        "2) *PART ... (nodes, elements, nsets/elsets, sections)\n"
        "   - All section assignments MUST be inside the *PART block\n"
        "   - If there is a *PART, there MUST be a matching *END PART\n"
        "3) *MATERIAL ... (all materials)\n"
        "4) *ASSEMBLY ... (instances, assembly-level sets/surfaces)\n"
        "   - If there is an *INSTANCE, there MUST be a matching *END INSTANCE\n"
        "   - If there is an *ASSEMBLY, there MUST be a matching *END ASSEMBLY\n"
        "5) *STEP ...\n"
        "   - *BOUNDARY must be inside the *STEP\n"
        "   - All loads (*CLOAD, *DLOAD, *DSLOAD) must be inside the *STEP\n"
        "   - Output requests (*OUTPUT, *NODE OUTPUT, *ELEMENT OUTPUT) must be inside the *STEP\n"
        "   - If there is a *STEP, there MUST be a matching *END STEP\n"
        "HARD RULES (must never be violated):\n"
        "   - The finite element mesh MUST contain at least 20 elements. "
        "     There is no upper limit on the number of elements.\n"
        "   - When requesting *ELEMENT OUTPUT, do NOT request output for any specific element set; "
        "     element output must be global/default only.\n"
        "   - Do NOT request *HISTORY OUTPUT under any circumstances.\n"
        "Before finalizing, validate that all referenced set names, material names, and section names exist "
        "and are used consistently."
    )
\end{Verbatim}
\end{systempromptbox}

\vspace{4pt}

\begin{userpromptbox}{User Prompt --- Rewrite Abaqus Input File}
\begin{Verbatim}[breaklines=true, fontsize=\small, breakanywhere=true]
 rewrite_user_prompt = (
        f"<Abaqus files>{str(Abaqusfiles)}</Abaqus files>\n"
        f"<error_logs>{error_logs}</error_logs>\n"
        f"<reviewer_analysis>{review_analysis}</reviewer_analysis>\n\n"
        f"<user_requirement>{user_requirement}</user_requirement>\n\n"
        "Please update the relevant Abaqus files to resolve the reported errors, ensuring that all modifications strictly adhere to the specified formats. Ensure all modifications adhere to user requirement."
    )   
\end{Verbatim}
\end{userpromptbox}

\subsection{Runner Agent Prompts}
\begin{systempromptbox}{System Prompt --- Local Execution Instructions}
\begin{Verbatim}[breaklines=true, fontsize=\small, breakanywhere=true]
    system_prompt = (
        "You are an expert in AbaqusAgent workflow analysis. "
        "The current AbaqusAgent setup supports only local Abaqus execution using an installed and licensed Abaqus software environment. "
        "Analyze the user requirement and return only 'local_run'. "
        "Do not return anything else."
    )

\end{Verbatim}
\end{systempromptbox}

\vspace{4pt}

\begin{userpromptbox}{User Prompt --- Local Execution Query}
\begin{Verbatim}[breaklines=true, fontsize=\small, breakanywhere=true]
    user_prompt = (
        f"User requirement: {user_requirement}\n\n"
        "Return only: 'local_run'."
    )
\end{Verbatim}
\end{userpromptbox}

\subsection{Reviewer Agent prompts}

\begin{systempromptbox}{System Prompt --- Error Diagnosis and Debugging}
\begin{Verbatim}[breaklines=true, fontsize=\small, breakanywhere=true]

REVIEWER_SYSTEM_PROMPT = (
    "You are an expert in Abaqus finite element modeling and Abaqus input (.inp) file debugging. "
    "Your task is to review the provided Abaqus error logs and diagnose the underlying issues. "
    "You will be provided with a similar case reference, which is a list of similar cases ordered by similarity. "
    "Use these references to understand the intended analysis and typical keyword patterns. "
    "When an error indicates that a specific Abaqus keyword, option, set name, surface name, element type, material name, or step procedure is undefined, "
    "your response must propose a fix that defines that exact missing item exactly as it appears in the error message (case-sensitive where applicable). "
    "Do not rename, reinterpret, or 'correct' the identifier; treat it literally and make the smallest change needed to define it. "
    "When the error indicates an unavailable procedure (e.g., a keyword not supported in Abaqus/Explicit such as *BUCKLE), "
    "propose an equivalent supported workflow using the same user intent (for example, eigenvalue buckling in Abaqus/Standard, "
    "or an Explicit dynamic alternative such as an imperfection + quasi-static compression), without changing the user-stated requirements unless unavoidable. "
    "Propose ideas to resolve the errors, but do not modify any files directly and do not output a complete rewritten .inp. "
    "Do not propose solutions that require changing geometry, loads, boundary conditions, material constants, units, or analysis targets stated in the user requirement; "
    "instead, prioritize fixes such as correcting keyword order, adding required data lines, defining missing *NSET/*ELSET/*SURFACE, "
    "fixing section assignments, adding amplitude/step controls, correcting node/element numbering, resolving duplicate IDs, "
    "fixing contact/rigid body definitions, and ensuring assembly/instance scoping is correct. "
    "The user will supply all relevant Abaqus files and the error logs; in the logs you will find the error content and the corresponding command or job context. "
    "Do not ask the user any questions. "
    "Your output must be a clear list of: (1) likely root cause(s), (2) the minimal keyword-level fix(es), and (3) a brief verification checklist "
    "to confirm the fix (e.g., 'check set exists at the correct scope', 'confirm step type supports the requested output')."
)

\end{Verbatim}
\end{systempromptbox}

\vspace{4pt}

\begin{userpromptbox}{User Prompt --- Abaqus Input File Revision}
\begin{Verbatim}[breaklines=true, fontsize=\small, breakanywhere=true]
    if history_text:
        reviewer_user_prompt = (
            f"<similar_case_reference>{tutorial_reference}</similar_case_reference>\n"
            f"<abaqusfiles>{str(Abaqusfiles)}</abaqusfiles>\n"
            f"<current_error_logs>{error_logs}</current_error_logs>\n"
            f"<history>\n{chr(10).join(history_text)}\n</history>\n\n"
            f"<user_requirement>{user_requirement}</user_requirement>\n\n"
            f"I have modified the files according to your previous suggestions. If the error persists, please provide further guidance. Make sure your suggestions adhere to user requirements and do not contradict it. Also, please consider the previous attempts and try a different approach."
        )
    else:
        reviewer_user_prompt = (
            f"<similar_case_reference>{tutorial_reference}</similar_case_reference>\n"
            f"<abaqusfiles>{str(Abaqusfiles)}</abaqusfiles>\n"
            f"<error_logs>{error_logs}</error_logs>\n"
            f"<user_requirement>{user_requirement}</user_requirement>\n"
            "Please review the error logs and provide guidance on how to resolve the reported errors. Make sure your suggestions adhere to user requirements and do not contradict it."
        )
\end{Verbatim}
\end{userpromptbox}

\vspace{8pt}
\section{User prompts structure in case studies}
Each benchmark case is defined by a natural-language prompt that specifies essential problem parameters, such as analysis type, geometry, material properties, boundary conditions, loading conditions, and requested outputs. Supplementary information is included as necessary to minimize ambiguity and enable more accurate agent reasoning, especially for complex problems.
\subsection{ Plate Penetration by a Projectile}

\begin{userrequirementbox}{User Requirement Format}
Perform a dynamic penetration analysis of a 2D axisymmetric circular plate impacted by a high speed projectile using Abaqus/Explicit. Create a 2D axisymmetric deformable plate with a diameter of 200 mm and a thickness of 10 mm. Define the plate material with elastic plastic behavior, density, strain rate sensitivity, and a damage or shear failure model with element deletion to allow penetration and hole formation. Do not constrain the nodes along the axis of symmetry in the radial direction so that material can expand during penetration. Fully constrain the outer edge of the plate. Model the projectile as an analytical rigid surface with a 20 mm diameter impacting tip and assign its mass to a reference node. Position the projectile tip slightly inside the contact region to avoid missed contact due to numerical precision. Define node based contact between the projectile and all plate nodes so that contact remains active even as elements fail and are removed. Define a Dynamic, Explicit step with duration based on projectile travel distance and initial velocity. Assign an initial velocity of 800 m/s to the projectile toward the plate center along the axis of symmetry. Extract displacement, stress, strain, damage, projectile velocity, and energy response to evaluate penetration behavior, material failure, and residual projectile velocity. Specifically, generate and display displacement contours and damage or element deletion evolution to visualize perforation.
\end{userrequirementbox}

\subsection{Buckling of Simply Supported Square Plate}
\begin{userrequirementbox}{User Requirement Format}
Perform a linear eigenvalue buckling analysis of a thin square simply supported plate under classical uniaxial in-plane compression. Model the plate as a flat shell midsurface of size 1.0 m × 1.0 m with thickness 10 mm. Use isotropic linear elastic material properties with Young’s modulus 200 GPa and Poisson’s ratio 0.3. Create a Linear Perturbation, Buckle step and extract the first four eigenmodes. Constrain U3 = 0 on all four edges to represent simple supports. Apply minimal in-plane constraints using corner nodes only to eliminate rigid body motion without breaking symmetry. Apply equal and opposite uniform in-plane compressive loads on the two opposite edges along the global x-direction, ensuring the loading is perfectly symmetric and introduces no moment or shear. Do not use edge-normal load definitions or surface pressure. Extract and display the buckling modes, where the first mode should show a single symmetric half-wave with maximum out-of-plane displacement at the plate center.
\end{userrequirementbox}

\subsection{Laminated Strip Under Three Point Bending}
\begin{userrequirementbox}{User Requirement Format}
Perform a static, linear elastic analysis of a 3D composite laminated strip to 
evaluate interlaminar response. Create a 3D deformable solid extrusion with a 
length of 300~mm, a width of 40~mm, and a total thickness of 10~mm. Define the 
material as a carbon fiber reinforced polymer (CFRP) with orthotropic elastic 
properties: $E_1 = 150$~GPa, $E_2 = E_3 = 10$~GPa, $\nu_{12} = \nu_{13} = 0.28$, 
$\nu_{23} = 0.40$, $G_{12} = G_{13} = 5$~GPa, and $G_{23} = 3.8$~GPa. Define a 
single Static, General step. Apply simple supports (constrained $U_2$) at two 
lines on the bottom face located 25~mm from each end. At the top center 
($x = 150$~mm), apply a downward concentrated force of 1500~N across the width 
through a reference point coupled to the loading line across the strip width. 
Extract the full body Displacement ($U$) and Stress ($S$) fields. Specifically, 
generate and display the full body displacement contours, full body longitudinal 
stress ($S_{11}$) contours, and full body transverse shear stress contours 
relevant to interlaminar response.
\end{userrequirementbox}

\subsection{Dynamic Buckling of X-Shaped Column with Contact}
\begin{userrequirementbox}{User Requirement Format}
Perform a transient dynamic analysis of an X-section column under impact. Create 
a 3D deformable solid extrusion with an X-shaped cross-section (web thickness 
10~mm, total width/height 100~mm) and a length of 1.5~m. Define the material as 
an elastoplastic steel ($E = 210$~GPa, $\nu = 0.3$, density $= 7850$~kg/m$^3$, 
yield stress $= 300$~MPa). Define a Dynamic, Explicit step. Fix the base of the 
column (fully constrained, ENCASTRE). Apply an initial velocity of 10~m/s to a 
rigid mass plate impacting the top of the column. Define General Contact to handle 
self-contact during buckling. Extract the full body Displacement ($U$) and Stress 
($S$) fields. Specifically, generate and display the full body displacement 
contours and full body Von Mises stress ($S$, Mises) contours at critical time 
increments, including the onset of collapse.
\end{userrequirementbox}

\subsection{Beam Under Moving Load
}
\begin{userrequirementbox}{User Requirement Format}
Perform a transient dynamic analysis of a simply supported beam subjected to a 
moving point load. Create a beam of length 5.0~m with a rectangular cross-section 
of 100~mm $\times$ 200~mm. Define the material as linear elastic steel with 
Young's modulus $E = 200$~GPa and Poisson's ratio $\nu = 0.30$. Use beam elements 
with appropriate section properties. Model simple supports by restraining vertical 
displacement at both ends, restraining axial displacement at one end only to 
remove rigid body motion, and allowing end rotation. Apply a vertical point load 
of 1000~N that moves from the left support to the right support over 2.0~s by 
defining the load location as a function of time. Use a transient dynamic analysis 
step. Extract the time-dependent displacement field, stress field, and beam 
rotation response to evaluate the moving load effect.
\end{userrequirementbox}

\end{document}